\pdfoutput=1

\documentclass[10pt,twocolumn,letterpaper]{article}

\usepackage{cvpr}
\usepackage{times}
\usepackage{graphicx}
\usepackage{amsmath}
\usepackage{amssymb}
\usepackage{float}
\usepackage[usenames, dvipsnames]{color}
\DeclareGraphicsExtensions{.pdf}
\usepackage[bottom]{footmisc}

\usepackage{algorithm2e}
\let\savedalgorithm\algorithm
\let\savedendalgorithm\endalgorithm

\usepackage{algorithm}
\usepackage{footnote}
\makesavenoteenv{tabular}
\usepackage{mathrsfs}

\def\T{{\!\top}}

\def\bx{ {\bf x } }

\def\bz{ {\bf z } }

\def\bw{ {\bf w } }

\def\bX{ {\bf X } }
\def\bS{ {\bf S } }
\def\bI{ {\bf I } }
\def\bA{ {\bf A } }

\def\bW{ {\bf W } }
\def\bD{ {\bf D } }

\def\bC{ {\bf C } }
\def\bZ{ {\bf Z } }
\def\bY{ {\bf Y } }
\def\bV{ {\bf V } }
\def\bB{ {\bf B } }

\newcommand{\tcb}[1]{\textcolor{blue}{#1}}
\newcommand{\tcr}[1]{\textcolor{red}{#1}}

\newcommand{\tcg}[1]{\textcolor{green}{#1}}

\def\etc{{\rm etc.}}

\cvprfinalcopy %

\begin{document}

\title{Less is more: zero-shot learning from online textual documents \\ with noise suppression\thanks{\color{blue}{Appearing in Proc.\ IEEE Conf.\ Computer Vision
and Pattern Recognition, 2016.} R. Qiao and L. Liu
  contribute equally to this work. Correspondence should be addressed to C. Shen.}
}

\author{Ruizhi Qiao, Lingqiao Liu, Chunhua Shen, Anton van den Hengel\\
School of Computer Science, The University of Adelaide, Australia\\
{\tt\small \{ruizhi.qiao,lingqiao.liu,chunhua.shen,anton.vandenhengel\}@adelaide.edu.au}
}

\maketitle
\thispagestyle{empty}

\begin{abstract}
Classifying a visual concept merely from its associated online textual source, such as a Wikipedia article, is an attractive research topic in zero-shot learning because it alleviates the burden of manually collecting semantic attributes. Several recent works have pursued this approach by exploring various ways of connecting the visual and text domains. This paper revisits this idea by stepping further to consider one important factor: the textual representation is usually too noisy for the zero-shot learning application. This consideration motivates us to design a simple-but-effective zero-shot learning method
capable of suppressing noise in the text.
More specifically, we propose an $l_{2,1}$-norm based objective function which can simultaneously suppress the noisy signal in the text and learn a function to match the text document and visual features. We also develop an optimization algorithm to efficiently solve the resulting problem. By conducting experiments on two large datasets, we demonstrate that the proposed method significantly outperforms the competing methods which rely on online information sources but without explicit noise suppression. We further make an in-depth analysis of the proposed method and provide insight as to what kind of information in documents is useful for zero-shot learning.

\end{abstract}

\tableofcontents
\newpage

\section{Introduction}
Unlike traditional object classification tasks in which the training  and test categories are identical, zero-shot learning aims to recognize objects from classes not seen at the training stage. It is recognized as an effective way for large scale visual classification since it alleviates the burden of collecting sufficient training data for every possible class. The key component ensuring the success of zero-shot learning is to find an intermediate semantic representation to bridge the gap between seen and unseen classes. In a nutshell, with this semantic representation we can first learn its connection with image features and then transfer this connection to unseen classes. So once the semantic representation of an unseen class is given, one can easily classify the image through the learned connection.

Attributes, which essentially represent the discriminative properties shared among both seen and unseen categories, have become the most popular semantic representation in zero-shot learning \cite{Farhadi09describingobjects, Ferrari_learningvisual07, TorresaniSzummerFitzgibbon10, Lampert09unseen, Yao11attributes}. Although the recent use of attributes has led to exciting advances in zero-shot learning \cite{Fu15manifold, Akata15output, Zhang2015ICCV}, the creation of attributes still relies on much human labour. This is inevitably discouraging since the motivation for zero-shot learning is to free large-scale recognition tasks from cumbersome annotation requirements.

To remedy this drawback and move towards the goal of fully automatic zero-shot learning, several recent works \cite{SocherNIPS2013, Frome2013NIPS_devise, NorouziConse13} have explored the possibility of using the easily accessed online information sources to create the intermediate semantic representation. One possible choice is to directly use  online textual documents, such as are found in Wikipedia, to build such a representation \cite{Elhoseiny2013ICCV, Ba2015ICCV}. This is promising because online text documents can be easily obtained and contain rich information about the object. To conduct zero-shot learning with textual documents, existing works \cite{Akata15output, Fu15manifold} develop various ways to measure the similarity between text and visual features. Our work is also based on this idea but we take a step further, however, to consider one additional important factor: the document representation is much more noisy than the human specified semantic representation and ignoring this fact will lead to inferior performance. For example, when the bag-of-words model is adopted as the document representation, the occurrence of every word in a document will trigger a signal in one dimension of the document representation. However, it is clear that most words in a document are not directly relevant for identifying the object category, thus it is necessary to design a noise suppression mechanism\footnote{This mechanism is closely related to feature selection
but is not exactly same. As will be discussed in the following sections,
the solution of our method does not discard the less relevant dimensions of the document representation but only suppress their impact for zero-shot learning.}  to down-weight the importance of the less relevant words for zero-shot learning.

Motivated by this consideration, we propose a zero-shot learning method which particularly caters for the need for noise suppression. More specifically, we proposed a simple-but-effective $l_{2,1}$-norm based objective function which simultaneously suppresses the noisy signal within text descriptions and learns a function to match the visual and text domains. We also develop an efficient optimization algorithm to solve this problem. By conducting experiments on two large scale zero-shot learning evaluation benchmarks, we demonstrate the benefit of the proposed noise suppression mechanism as well as its superior performance over other zero-shot learning methods which also rely on online information sources. In addition, we also conduct an in-depth analysis of the proposed method which provides an insight as to what kinds of information within a document are useful for zero-shot learning.

\section{Related work}
\label{sec: rel_work}
Most zero-shot learning approaches rely on human specified attributes. As one of the earliest attempt in zero-shot learning,  Lampert \etal \cite{Lampert09unseen} adopted a set of attributes obtained from a psychology study. By learning probabilistic predictors of those attributes, they developed a framework to estimate the posterior of the test class. Later, a number of works has been proposed to improve the way of learning the connnection between attributes and object categories. For example, the work in \cite{Jayaraman14NIPS} addresses unreliability of attributes by exploring the idea of random forest. The work in \cite{Akata13label} turns the zero-shot learning into a cross-domain matching problem and they propose to learn a matching function to compare the attribute and the image feature. Following the same idea, Romera \cite{Romera2015ZSL} proposes a simpler but more effective objective function to learn the matching function. Zhang \etal \cite{Zhang2015ICCV} advocates the benefits of using attribute-attribute relationship, called semantic similarity, as the intermediate semantic representation and they learn a function to match the image features with the semantic similarity.

To go beyond the human specified attributes, recent works also explore the use of other form of semantic representations which can be easily obtained \cite{Mensink14CVPRCOSTA,Akata15output,Frome2013NIPS_devise,Fu15manifold}. For example, the co-ocurrence statistic of words has been explored in \cite{Mensink14CVPRCOSTA,Akata15output} to capture the semantic relateness of two concepts. The distributed word representation e.g. word2vec, has been utilized as a substitution of attributes \cite{Frome2013NIPS_devise} and more recently the word2vec representation has been shown to be complementary to the human specified attributes \cite{Fu15manifold}.

Another information source for creating the semantic representation is the online textual document, such as Wikipedia articles. In an earlier work, Berg \etal \cite{Berg10disc} attempts to discover attribute representation from a noisy web source by ranking the visualness scores of attribute candidates. Rohrbach \etal \cite{Rohrbach13NIPS, Rohrbach10CVPR}
mines semantic relatedness for attribute-class association
from different internet sources. More recent works \cite{Elhoseiny2013ICCV, Ba2015ICCV} directly learn a function to measure the compatibiliy between documents and visual features. However, compared with the state-of-the-art zero-shot learning methods, their performance seems to be disappointing even though some advanced technologies, such as deep learning, has been applied \cite{Ba2015ICCV}.

\begin{figure*}[t]
\begin{center}
\label{fig:overview}
   \includegraphics[width=1\linewidth]{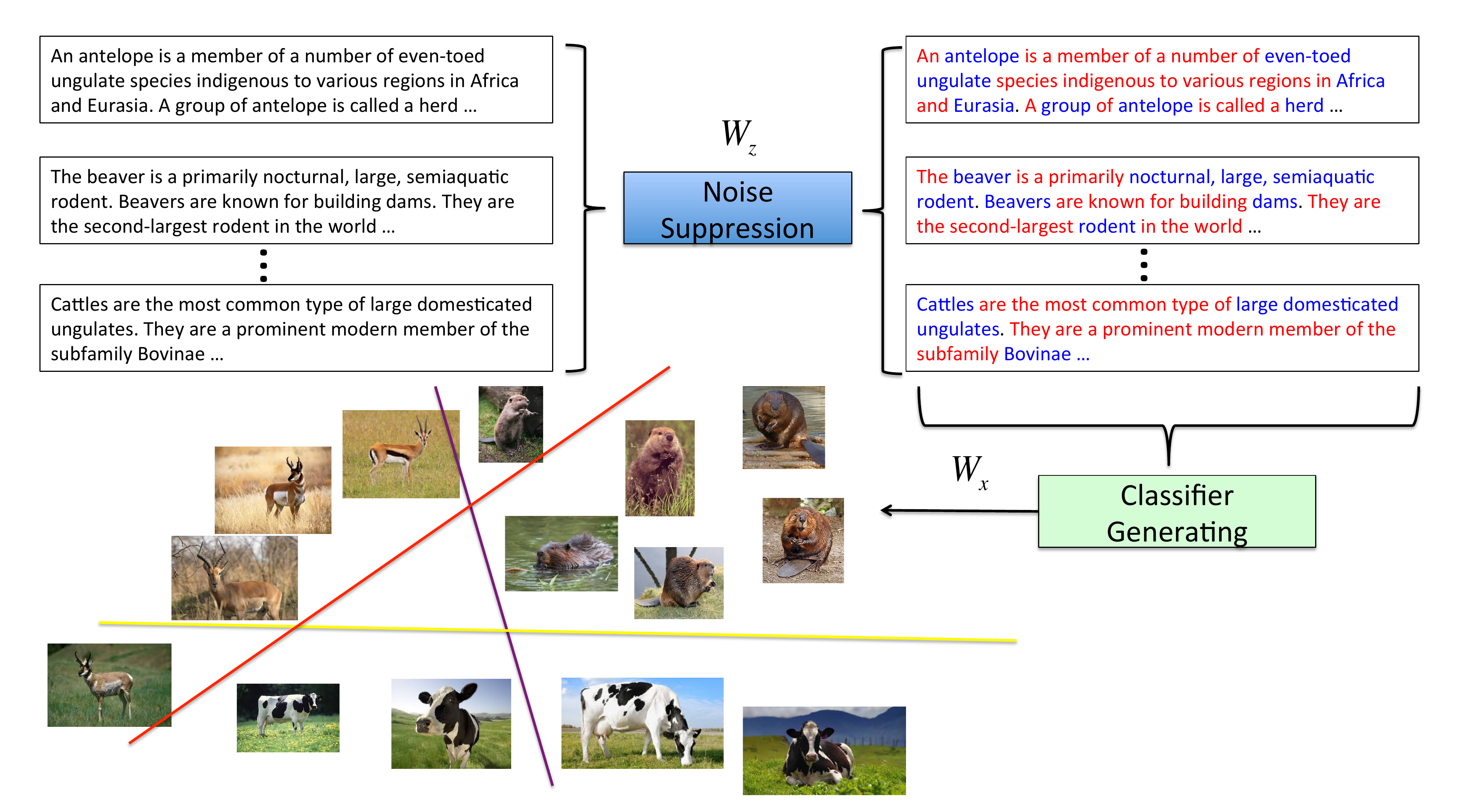}
\end{center}
   \caption{Overview of our zero-shot learning approach. The text representations are processed by the noise suppression mechanism to generate a classifier to detect relevant images and the noisy components of text representations are suppressed to gain better performance.}
\label{fig:overview}
\end{figure*}

\section{Our Approach}

\subsection{Overview}
The overview of our method is depicted in Figure \ref{fig:overview}. It starts with a raw document representation which is simply a binarized histogram of words. This representation is fed into our zero-shot learning algorithm to generate a classifier to detect relevant images. In the process of generating this classifier, the noise suppression regularizer in our method will automatically suppress the impact of less relevant words (illustrated as the red words in Figure \ref{fig:overview}).

\subsection{Text representation}
\label{sec: text}
We extract our text representation based on a simple bag-of-words model. We start by a preprocessing step of tokenizing the words and removing stop words and punctuations. Then a histogram of the remaining word occurrences is calculated and is subsequently binarized as the text representation. In other words, once a word appears in a document, its corresponding dimension within the text representation is set to ``1''. One more commonly used choice for the text representation is based on TF-IDF as in \cite{Elhoseiny2013ICCV, Ba2015ICCV}. However, we find it produces  worse performance \footnote{Using TD-IDF is about 7\% and 5\% inferior to binarized representations on AwA and CUB, respectively.} than directly using the binarized representation. This is probably because the weighting calculated of TF-IDF is not suitable for our zero-shot learning although it is considered to be less noisy for applications like document classification. In the binarized  histogram we essentialy treat each word in a document equally and this inevitably introduces a lot of noisy signals. However, thanks to our noise suppressing zero-shot learning algorithm, we can substantially down-weight the less relevant words and achieve good performance even with a noisy document representation.

\subsection{Learning to match text and visual features}

We first formally define our problem and introduce the notation used in the following sections. At the training stage, both image features and document descriptions for $C$ seen categories are available. Let $\bX \in \mathbb{R}^{d\times N}$ denote the image features of $N$ training examples and $\bZ \in \{0,1\}^{\hat{d} \times C}$ the aforementioned document representations for $C$ seen classes, where $\hat{d}$ and $d$ are the dimensionality of the document representation and the image features respectively. We also define $\bY \in \{0,1\}^{N \times C}$ as the indicator matrix for the $C$ seen classes. Each row of $\bY$ has a unique ``1'' indicating its corresponding class label. At the test stage, the document representations of the $\hat{C}$ unseen classes are given and our task is to assign $\hat{C}$ unseen class labels to the test images.

\subsection{Formulation}
 Our method is inspired by a recently proposed zero-shot learning approach \cite{Romera2015ZSL} which has demonstrated impressive performance despite a very simple  learning process. More specifically, it learns a matrix $\bV$ which optimizes the following objective function.
\begin{align}\label{eq:ESL}
\min_{\bV}  \|\bX^\T \bV \bS - \bY\|_F^2 + \lambda \| \bV \bS \|_F^2 + \gamma \| \bX^\T \bV \|_F^2 + \lambda \gamma \|\bV\|_F^2
\end{align}where $\bS$ denotes the semantic attribute matrix and it can be either a binary matrix or a real value matrix.
The scalars $\gamma$ and $\lambda$ are weights controlling the prominence of the various terms.
The underlying idea of this algorithm can be understood as follows. If the task is to classify $\bX$ into $C$ categories, we can simply learn a linear classifier
by fitting to
$\bY$, that is, $\min_{\bW} \|\bX^\T \bW - \bY\|_F^2$. However, in this case $\bW$ cannot be transferred to unseen classes. Thus we further impose that $\bW = \bV\bS$. In other words, the classifier of a class is generated from its attributes. With this requirement, the classifier of an unseen class can be easily obtained and utilized to predict the category of a test image. Similarly, we can also treat $\bX^\T \bV$ as the classifier operated on the attributes $\bS$. The above understanding naturally gives rise to the regularization terms $\lambda\|\bV \bS\|_F^2$ and $\gamma\|\bX^\T \bV\|_F^2$ which play the same role of the Frobenius norm regularizer as commonly introduced in multi-class classification or regression.

Since our document representation can also be seen as an attribute vector, the method in \cite{Romera2015ZSL} can be readily applied to our problem by simply setting $\bS = \bZ$. However, this naive solution ignores an important fact that the document representation is much more noisy than the human specified attribute vectors. To handle this issue, we introduce a noise suppression mechanism into Eq. (\ref{eq:ESL}). More specifically, we first decompose $\bV$ into two terms:
\begin{align}
	\bV = \mathbf{W_x}^\T \mathbf{W_z},
\end{align}where $\mathbf{W_x} \in \mathbb{R}^{m \times d}$ and  $\mathbf{W_z} \in \mathbb{R}^{m \times \hat{d}}$. These two matrices will play different roles in our method. $\mathbf{W_z}$ is used to suppress the noisy components of $\bZ$ and transform $\bZ$ into a $m \times C$ intermediate representation. $\mathbf{W_x}$ is used to generate the image classifier from the noise-suppressed intermediate representation. Thus, two different regularization terms are imposed to suit these two different roles. The first term is the
$l_{2,1}$-norm of $\mathbf{W_z}^\T$ which achieves the noise suppression effect. The second term is the Frobenius norm of $\mathbf{W_x}^{\T} \mathbf{W_z} \bZ$ which is similar to the $\lambda\|\bV \bS\|_F^2$ term in Eq. (\ref{eq:ESL}). The formulation of our method is expressed as follows:
 \begin{align}
            \label{EQ: formulation}
		\min_{\bW_\bx, \bW_\bz} \; &  L(\bW_\bx, \bW_\bz) + \lambda_1 \| \bW_\bx^\T \bW_\bz \bZ\|_F^2 + \lambda_2 \|\bW_\bz^\T\|_{2,1} ,
		\\
		& L(\bW_\bx, \bW_\bz) = \| \bX^\T \bW_\bx^\T \bW_\bz \bZ - \bY\|^2_F. \notag
\end{align}

The $l_{2,1}$-norm is defined as $\|\bW_\bz^T\|_{2,1} = \sum_{i=1}^{\hat{d}} \|\bw_\bz^i\|_2$, where $\bw_\bz^i$ denotes the $i$-th column of $\bW_\bz$. It is known that the $l_{2,1}$-norm will encourage the column vectors of $\mathbf{W_z}$ to have few large values, which means that the impact of noisy dimensions of $\bZ$ will be substantially suppressed or even completely eliminated. In fact, if $\lambda_2$ becomes sufficient large, it achieves the effect of feature selection on the document representation. However, by cross-validating $\lambda_1$ and $\lambda_2$, our method does not lead to an exactly sparse solution as it seems that the algorithm prefers to keep the majority of the dimensions in $\bZ$ for zero-shot learning. This is probably due to the joint regularization effect of $\| \bW_\bx^\T \bW_\bz \bZ\|_F^2$ or the fact that dimensions corresponding to lower values of $\|\bw_\bz^i\|_2$ are still useful for zero-shot learning.
Therefore we consider the use of the $l_{2,1}$-norm here as a noise suppression mechanism rather than a feature selection mechanism. We drop out the other regularization terms in Eq. (\ref{eq:ESL}) since we find them have little impact on  performance.

Similar to \cite{Romera2015ZSL}, once $\bV$, in our case $\bV = \mathbf{W_x}^\T \mathbf{W_z}$, is learned, we can infer the class label of a test image $\bx$  using the following rule:
\begin{align}
	c^* = \max_{c} \, \bx^\T \bW_\bx^\T \bW_\bz \bz_c,
\end{align}where $\bz_c$ is the document representation for the $c$-th candidate test class.

\subsection{Optimization}

Eq. \eqref{EQ: formulation} is convex for $\bW_\bx$ and $\bW_\bz$ individually but not convex for both of them. Therefore we can solve it using an
alternating
method, that is, we first fix $\mathbf{W_x}$ and solve for $\mathbf{W_z}$; then fix $\mathbf{W_z}$ and solve for $\mathbf{W_x}$.

\noindent \textit{ (1) Fix $\mathbf{W_x}$ and solve for $\mathbf{W_z}$}:
\def\ADot{ { $\bf \cdot$ } }%

\setcounter{AlgoLine}{1}
\begin{algorithm}[ht]
\caption{ Fix $\mathbf{W_x}$ and solve $\mathbf{W_z}$}
\centering   
{
   \begin{minipage}[]{0.94 \linewidth}
    \KwIn{ $\bW_\bx$;
    $\bX $ of seen classes; $\bZ$ of seen classes; $\lambda_1$ and $\lambda_2$; maximum number of iterations $\tau$.
     
    }   
   { Initialize $\bD^0$ as identity matrix $\bI \in \mathbb{R}^{\hat{d} \times \hat{d}}$.
   }
   
   \For { $t = 1 \cdots \tau$  }
   {
    \ADot
    		Solve Sylvester equation \eqref{EQ: Sylvester}  for $\bW_\bz^t$ with $\bD^{t-1}$.

    \ADot
        Update the diagonal matrix $\bD^t$ with its $i$-diagonal element as $1 / (2 ||(\bw_\bz^i)^{(t)}||_2)$, where $(\bw_\bz^i)^{(t)}$ is the $i$-th column of $\bW_\bz^t$.

     \If{\it Converges}
      {
         \ADot
            Break.
      }

   }

\KwOut{
     $\bW_\bz$.
}
\end{minipage}
}
\label{ALG:Wz}
\end{algorithm}
 This sub-problem is a regression problem with $l_{2,1}$-norm regularization. Nie \etal \cite{Nie2010NIPS} proposes an iterative framework to efficiently solve it. It has been shown that the original problem is equivalent to sequentially solving the following problem  until convergence%
	 \begin{align}
            \label{EQ: iter_W}
		\min_{\bW_\bz, \bD} \; L(\bW_\bx, \bW_\bz) + \lambda_1 \| \bW_\bx^\T \bW_\bz \bZ\|_F^2 + \\ \nonumber \lambda_2 Tr(\bW_\bz \bD^t \bW_\bz^\T),
        \end{align}
where $\bD^t$ is a diagonal matrix whose $i$-th diagonal element is $1 / (2 \|(\mathbf{ w}_z^i)^{(t-1)}\|_2)$\footnote {In practice, we relax $1 / (2 ||\bw_\bz^i||_2)$ to $1 / (2 \sqrt{{\bw_\bz^i}^\T {\bw_\bz^i} + \sigma})$, $\sigma \rightarrow 0$, as the $i$-th diagonal element to avoid the case of zero columns, and the $l_{2,1}$ norm is therefore approximated by $\sum_{i=1}^{\hat{d}} \sqrt{{\bw_\bz^i}^\T {\bw_\bz^i} + \sigma}$. It has been proved in \cite{Nie2010NIPS} that this approximation guarantees the convergence and the result approaches to that of $l_{2,1}$-norm as $\sigma \rightarrow 0$ .} at the $t$-th iteration, where $(\mathbf{ w}_z^i)^{(t-1)}$ is the $i$-th column of the optimal $\mathbf{W_z}$ solved at the $(t-1)$-th iteration.
The problem in Eq. (\ref{EQ: iter_W}) further reduces to a Sylvester equation of $\bW_\bz$

	 \begin{align}
            \label{EQ: Sylvester}
		& \bA \bW_\bz  + \bW_\bz \bB = \bC,
		\\
		& \bA = \lambda_2 (\bW_\bx \bX \bX^\T \bW_\bx^\T + \lambda_1 \bW_\bx \bW_\bx^\T )^{-1}, \notag
		\\
		& \bB = \bZ \bZ ^\T (\bD)^{-1}, \notag
		\\
		& \bC = \frac{1}{\lambda_2} \bA \bW_\bx \bX \bY \bZ^\T  (\bD)^{-1}. \notag
        \end{align}
The Sylvester equation has a unique solution if and only if $\bA$ and $-\bB$ do not share any eigenvalues. Many state-of-the-art toolboxes are able to solve it efficiently. In our setting, since both $\bA$ and $\bB$ are positive definite, $\bA$ has only positive eigenvalues and $-\bB$ has only negative eigenvalues. Therefore Eq.~\eqref{EQ: Sylvester} has a unique solution. In summary, the sub-problem of fixing $\mathbf{W_x}$ to solve $\mathbf{W_z}$ can be solved via the algorithm listed in Algorithm \ref{ALG:Wz}.

\noindent \textit{ (2) Fix $\mathbf{W_z}$ and solve for $\mathbf{W_x}$}:
\def\ADot{ { $\bf \cdot$ } }%

\setcounter{AlgoLine}{1}
\begin{algorithm}[ht]
\caption{ Alternating algorithm for solving Eq.~\eqref{EQ: formulation}}
\centering   
{
   \begin{minipage}[]{0.94 \linewidth}
    \KwIn{
     $\bX$ of seen classes;  $\bZ$ of seen classes; $\lambda_1$ and $\lambda_2$; maximum number of iterations $\tau$.
     
    }   
   { Initialize $\bW_\bx^0$ with Gaussian distribution.
   }
   
   \For { $t = 1 \cdots \tau$  }
   {
    \ADot
    		Solve \eqref{EQ: iter_W}  iteratively for $\bW_\bz^t$ with $\bW_\bx^{t-1}$ according to Algorithm~\ref{ALG:Wz}.

    \ADot
        Solve \eqref{EQ: iter_Wx} for $\bW_\bx^t$ with $\bW_\bz^{t}$.

     \If{\it Converges}
      {
         \ADot
            Break.
      }

   }

\KwOut{
     $\bW_\bx$, $\bW_\bz$.
}
\end{minipage}
}
\label{ALG:all}
\end{algorithm}
This sub-problem is a conventional least squares minimization problem which has the following closed-form solution

	 \begin{align}
            \label{EQ: iter_Wx}
		\bW_\bx^\T = (\bX \bX^\T + \lambda_1 \bI)^{-1} \bX \bY \bZ^\T \bW_\bz^\T (\bW_\bz \bZ \bZ^\T \bW_\bz^\T)^{-1}.
        \end{align}

By alternating between the above two matrices, the overall alternating optimization algorithm for Eq. \eqref{EQ: formulation} is listed in Algorithm \ref{ALG:all}.

\section{Experiments}
\label{sec: exp}

We divide our experiments into two parts. In the first part we evaluate the proposed method and compare it against both of the methods utilizing online textual sources and human-specified semantic attributes. In the second part we analyse in-depth the noise suppression effect of the proposed method and provide insight into what kind of information in a document is useful for zero-shot learning.
\subsection{Experimental setting}
\label{sec: setting}
{\bf Datasets:} We test our approach on two widely used benchmarks for attribute learning and zero-shot learning: Animals with Attributes \cite{Lampert09unseen} (AwA) and Caltech-UCSD birds-200-2011 \cite{CUB_200_2011} (CUB-200-2011). AwA consists of 30,475 images of 50 mammals classes with 85 attributes including color, skin texture, body size, body part, affordance, food source, habitat, and behaviour. CUB-200-2011 contains 11,788 images of 200 categories of bird subspecies with 312 fine-grained attributes such as color/shape/texture of body parts. We follow the train/test split according to \cite{Lampert09unseen}  and \cite{CUB_200_2011}, where 10 and 50 testing classes are treated as unseen for AwA and CUB-200-2011, respectively.

{\bf Textual document sources:} We extract the text representation according to scheme introduced in Section~\ref{sec: text}. The raw textual sources are collected from Wikipedia articles describing each of the  categories. When constructing the vocabulary, we use the articles of seen classes only. The dimensionality of the text representation is 3506 for AwA and 6815 for CUB-200-2011, respectively.

{\bf Image features:} To make fair a comparison, two types of image features, the low-level features in \cite{Rohrbach10CVPR} and the fully connected layer activations from the ``imagenet-vgg-verydeep-19'' \cite{Simonyan14VGG} CNN are used in our experiments.

{\bf Implementation details:} The Sylvester equation in Eq.~\eqref{EQ: Sylvester} is solved by a MATLAB built-in function, which takes only around 5 seconds on an Intel Core i7 CPU at 3.40GHz. The number of rows of matrices $\bW_\bx$ and $\bW_\bz$ is equal to the number of seen classes. We choose the hyper-parameters with a five-fold cross-validation on the seen classes, where 20\% (5 for AwA and 30 for CUB-200-2011) of the seen classes are held out for validation and the remaining seen classes are used for training. The hyper-parameters are tuned within the range of all cases of $10^b$, where $b=\{ -2, -1, \cdots, 5, 6 \}$. Once the hyper-parameters are selected, we use all seen classes to train the final model. All of our reported results are averaged over 10 trials.

\subsection{Performance evaluation}

\begin{table}[h]
\begin{center}
\begin{tabular}{ l|c|c }
\hline
Method &  Top-1 Acc  &  Top-5 Acc\\
\hline\hline
Ba \cite{Ba2015ICCV} (BCE) & 1  & 17.6\\
Ba \cite{Ba2015ICCV} (Hinge) & 0.6 & 18.2 \\
Ba \cite{Ba2015ICCV} (Euclidean) & 12 & 42.8\\
ESZSL \cite{Romera2015ZSL}  & 23.80 & 59.90\\
Ours  & $29.00 \pm 0.28$ & $61.76 \pm 0.22$\\
\hline
\end{tabular}
\end{center}
\caption{Zero-shot learning classification results on CUB-200-2011, measured by top 1 and top 5 accuracy. 3 different loss functions are used in \cite{Ba2015ICCV} for their CNN structure: binary cross entropy (BCE), hinge loss (Hinge), and Euclidean distance (Euclidean). All methods in this table use the same text sources from Wikipedia.}
\label{tab: zsl-low-CUB}
\end{table}

\begin{table}[h]
\begin{center}
\begin{tabular}{ l|c }
\hline
Method & Mean Accuracy \\
\hline\hline
Rohrbach \cite{Rohrbach10CVPR} (Wikipedia) & 19.7 \\
Rohrbach \cite{Rohrbach10CVPR} (WordNet) & 17.8 \\
Rohrbach \cite{Rohrbach10CVPR} (Yahoo Web) & 19.5 \\
Rohrbach \cite{Rohrbach10CVPR} (Yahoo Img) & 23.6 \\
Rohrbach \cite{Rohrbach10CVPR} (Flickr Img) & 22.9 \\
ESZSL \cite{Romera2015ZSL} (Wikipedia)  & 24.82\\
Ours (Wikipedia) & $29.12 \pm 0.07$ \\
\hline
\end{tabular}
\end{center}
\caption{Zero-shot learning classification results of AwA, measured by mean accuracy. In \cite{Rohrbach10CVPR}, the approach mines attributes names from WordNet and additionally mines class-attribute from online sources of Wikipedia, WordNet, Yahoo, and Flickr. All methods in this table use the same low-level features in \cite{Rohrbach10CVPR}.}
\label{tab: zsl-low}
\end{table}

We first compare our method against \cite{Ba2015ICCV} and \cite{Rohrbach10CVPR}. The former is most relevant to our work in the sense that it learns a mapping to match images and textual documents. The work in \cite{Rohrbach10CVPR} is a comprehensive comparison study of various information sources for zero-shot learning. Besides these two method, we also treat $\bS=\bZ$ in Eq.~\eqref{eq:ESL}, and apply the ESZSL method in \cite{Romera2015ZSL} to our zero-shot learning problem. To make a fair comparison, we use the same low-level features in \cite{Rohrbach10CVPR} when comparing with it and then use the ``imagenet-vgg-verydeep-19'' to compare with \cite{Ba2015ICCV}. The comparison results are given in Table~\ref{tab: zsl-low-CUB} and Table~\ref{tab: zsl-low}. As can be seen in Table~\ref{tab: zsl-low-CUB}, the proposed method significantly outperforms the methods in \cite{Ba2015ICCV}, although they have used a more complicated deep learning framework. Also, we find that our baseline ESZSL achieves good performance. However, it is still 5\% inferior to our approach, which clearly demonstrates the advantage of the noise suppression mechanism introduced in this paper. The results in Table~\ref{tab: zsl-low} further show that our method is superior over other approaches which rely on automatically mined information from the web. Again, our method achieves a significant improvement (more than 4\%)  over ESZSL.

\begin{table}[h]
\begin{center}
\begin{tabular}{ l|c|c }
\hline
Method/Dataset & AwA & CUB\\
\hline\hline
Rohrbach \cite{Rohrbach13NIPS}  & 42.7 & \\
Jayaraman \cite{Jayaraman14NIPS}  & 43.01 &  \\
Mensink\cite{Mensink14CVPRCOSTA} &  & 14.4\\
Akata \cite{Akata13label}  & 43.5 & 18.0\\
\hline
\hline
Lampert \cite{Lampert14pami} (attr. real)  & 57.5 & \\
Deng \cite{deng2014large} (hierarchy) & 44.2 & \\
ESZSL \cite{Romera2015ZSL} (attr. bin)  & 62.85 &  \\
Akata \cite{Akata15output} (Word2Vec)  & 51.2 & 28.4\\
Akata \cite{Akata15output} (GloVe)  & 58.8 & 24.2\\
Akata \cite{Akata15output} (WordNet)  & 51.2 & 20.6\\
Akata \cite{Akata15output} (attr. bin)  & 52.0 & 37.8\\
Akata \cite{Akata15output} (attr. real)  & 66.7 & 50.1\\
Fu \cite{Fu15manifold} (attr. \& words) & 66.0 & \\
Zhang \cite{Zhang2015ICCV} (attr. real) & 76.33 & 30.41 \\
ESZSL \cite{Romera2015ZSL} (Wikipedia)  & 58.53 & 23.80  \\
Ours (Wikipedia)  & $66.46 \pm 0.42$ & $29.00 \pm 0.28$ \\
\hline
\end{tabular}
\end{center}
\caption{Zero-shot learning classification results on AwA and CUB-200-2011. Blank spaces indicate these methods are not tested on the corresponding datasets. Contents in braces indicate the semantic sources which these methods use for zero-shot learning. Methods in the upper part of the table use low-level features and the remaining methods in the lower part use deep CNN features. }
\label{tab: zsl-all}
\end{table}
We now let our work compete with other state-of-the-art approaches on zero-shot learning, even though some of them are not based on online information sources. The results are summarized in Table~\ref{tab: zsl-all}. Results \cite{Rohrbach13NIPS, Jayaraman14NIPS, Mensink14CVPRCOSTA, Akata13label} listed in the upper part of the table utilize hand-crafted features and not surprisingly their performance is much inferior to that of the proposed method. The lower part of Table~\ref{tab: zsl-all} are methods with visual features extracted from a pre-trained  CNN and thus are more comparable to our method. In this setting, we find that our method is comparable to most of the state-of-the-art results on AwA and results better than ours are all obtained from the methods using cleaner human defined attributes. The work in \cite{Akata15output} evaluates various semantic representations such as Word2Vec embedding, GloVe word co-occurrence from Wikipedia sources, taxonomy embedding inferred from WordNet Hierarchy, and pre-defined binary and real-valued attributes. Our approach outperforms all methods that use online text sources. This shows that although online text sources provide transferable semantic representations, their discriminative ability is affected by the inherent noise and our method is better at handling the noisy information source for zero-shot learning.

Similar results are observed on the CUB-200-2011 dataset. Our approach again outperforms the methods using online sources and those methods that beat ours are all based on human specified fine-grained attributes. Note that many of the bird categories in CUB-200-2011 have very subtle differences which may not be well captured in Wikipedia articles. However, better performance may be expected by using a higher quality text corpus, such as  bird watching articles.

\subsection{In-depth analysis of the proposed method}
\begin{figure*}[t]

\begin{center}

   \includegraphics[width=1\linewidth]{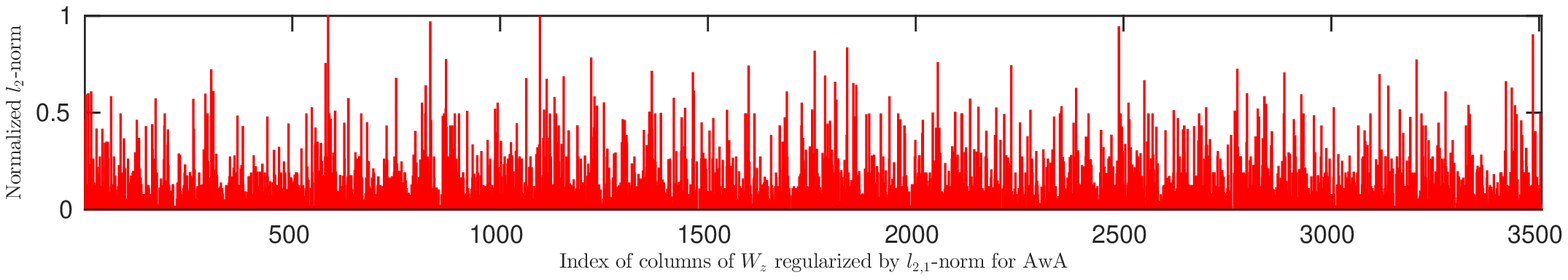}
   \includegraphics[width=1\linewidth]{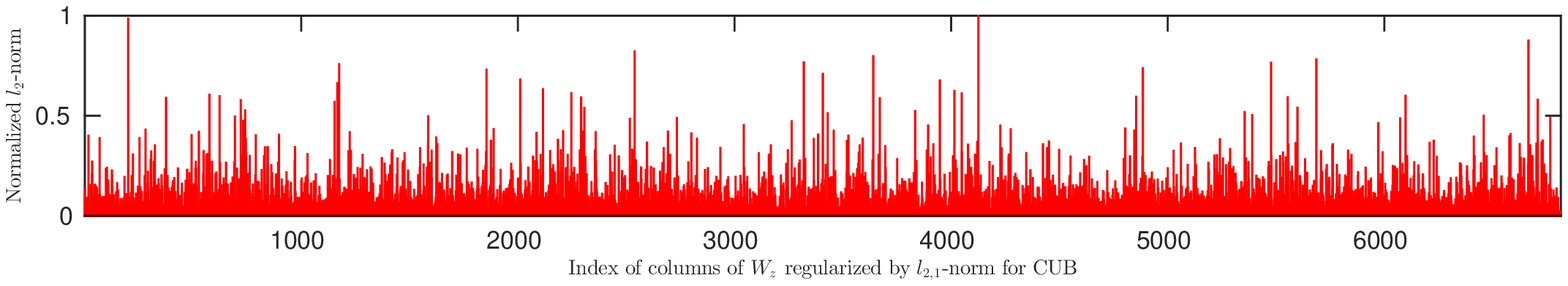}
   \includegraphics[width=1\linewidth]{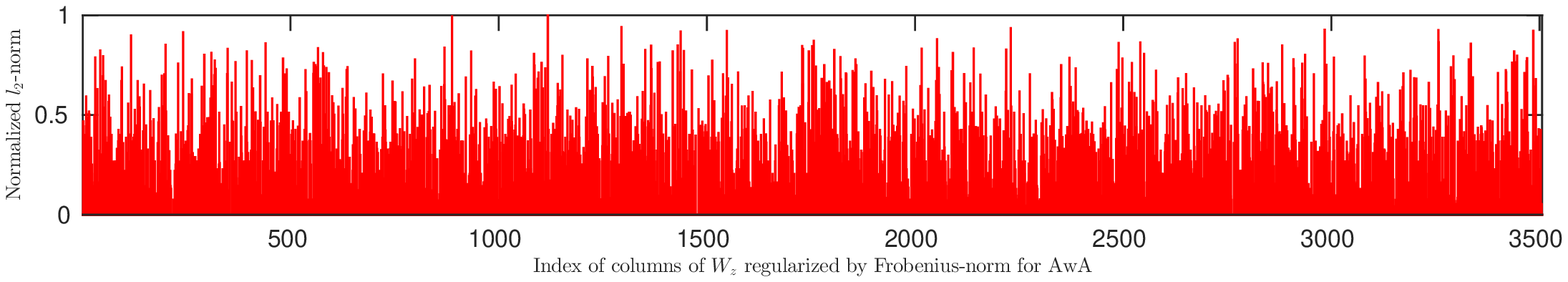}
   \includegraphics[width=1\linewidth]{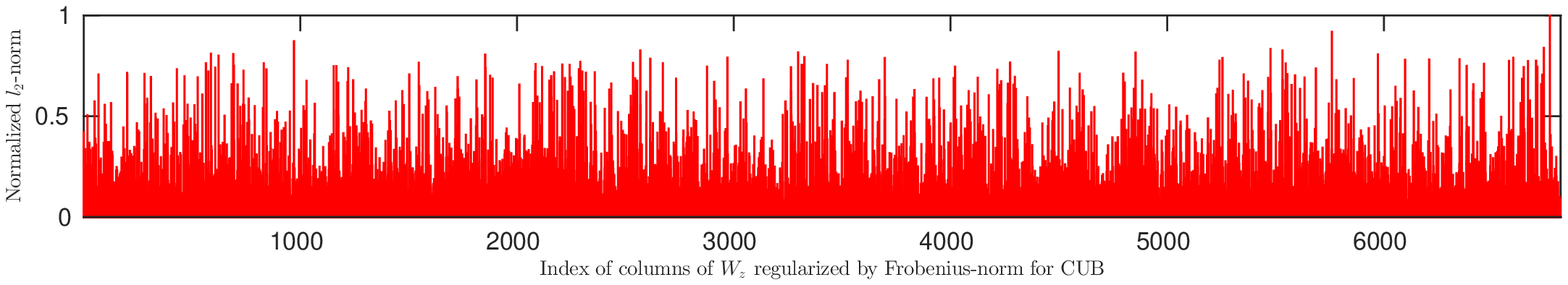}
\end{center}
   \caption{The two subfigures at the top show column-wise $l_2$-norms of $\bW_\bz$ learned with $l_{2,1}$-norm regularization. The two subfigures at the bottom show column-wise $l_2$-norms of $\bW_\bz$ learned with Frobenius-norm regularization. }
\label{fig: norm}
\end{figure*}
\label{sec: analysis}
In this section we provide an in-depth analysis of the proposed method by examining its noise suppression mechanism and the words that are most discriminative in the view of our method.
\subsubsection{Effectiveness of the noise suppression method}

In our method, the $l_{2,1}$-norm is expected to allow only a few dimensions of the document representation to have large values. The importance of each individual dimension of the document representation can therefore be measured by the $l_2$-norm of each column of learned $\bW_\bZ$ (we call it the importance weight in the following). We visualize this measurement for each dimension of the document representation in the top two subfigures in Figure~\ref{fig: norm}. As can be seen, most of the importance weights are not exactly zero as one might expect given that the $l_{2,1}$-norm is applied. In fact, there are only 702 zero columns (out of 3506) for AwA and 949 (out of 6815) for CUB-200-2011. As also mentioned in Section~\ref{EQ: formulation}, this is probably because of the joint regularization effect of $|| \bW_\bx^\T \bW_\bz \bZ||_F^2$ in Eq.~\eqref{EQ: formulation} and/or because by cross-validation most dimensions are still identified as being useful although their weighting should be very low. The second postulate might be supported by the observation that poorer performance will be obtained if we manually remove the dimensions which have low importance weights.

Although our formulation does not achieve the feature selection effect, it does only assign large importance weights to a small number of  dimensions. To visually compare its effect, we replace the $l_{2,1}$-norm and with the Frobenius norm and carry out our learning algorithm again. The resulting importance weights are shown in the two subfigures at the bottom of Figure~\ref{fig: norm}. As can be seen, large importance weights appear in more dimensions in this case. This observation verifies the noise suppression effect of the regularizer introduced in Eq.~\eqref{EQ: formulation} and explains the superior performance of our method over other text-based zero-shot learning approaches.

\subsubsection{Understanding the important dimensions of the document representation}

Since each individual dimension of the textual document representation corresponds to an unique word, we can visualize the dimensions/words with large importance weights for better understanding our zero-shot learning algorithm. Table~\ref{tab: discAttrAwA} lists at most 15 top scored words for 15 out of 40 seen classes in AwA and we could make several observations from it: (1) even though the document representations are extremely noisy, most of the top-ranked words are semantically meaningful to describe discriminative properties of a category (an animal in this case), such as body parts, habitat, behaviour, affordance, taxonomy, and environment. In fact, we find many top weighted words are consistent with some of the human specified attributes in AwA. (2) Many top-ranked words are not explicitly ``visualizable'' but they imply visual information of a category. For example,  the abstract concept ``ruminant'' implicitly tells that the creature with this property is "deer-like" or "cattle-like" and builds a visual connection between antelope and deer in Table~\ref{tab: discAttrAwA}. This observation has also been made in the literature \cite{Lampert09unseen, Lampert14pami, Osherson91default, Berg10disc, shao2015deepattribute}. (3) Interestingly, we also notice that although some concepts are not commonly considered as attributes, they exhibit large importance weight as inferred by our algorithm. By taking a close examination, we categorize these words into two types. The first (labelled green in Table~\ref{tab: discAttrAwA}) are some concepts that are more likely to co-occur with meaningful attributes. For example, the word ``stomach'' is only shared by antelope and deer in Table~\ref{tab: discAttrAwA}, despite its existence in all mammals. This is probably because ``stomach'' is more likely to be co-occurred with ``ruminant'', a discriminative property of ruminant animals. Another type of words (labelled red in Table~\ref{tab: discAttrAwA}) are not sufficiently meaningful for human interpreter.  For example, ``belong'' and ``general'' are assigned with high importance weight for all cetaceans (blue whale, dolphin, killer whale \etc) and rodents (mouse, rabbit, hamster \etc), respectively. We suspect the reason is due to the dataset bias of documents. For example, documents of similar categories may be edited by authors from the same background who prefer a certain word choice. In sum, we find most of the top ranked words carry weak information by their own, but it seems that using them collaboratively produces impressive discriminative power for zero-shot learning.

\begin{table*}[ht]
\begin{center}
  \small
\begin{tabular}{ l|c }
\hline
Class & Top Ranked Words/Dimensions \\
\hline\hline
Antelope & \tcb{antler}, \tcb{woodland}, \tcb{fight}, \tcg{stomach}, \tcb{spike}, \tcb{antelope}, \tcb{escape}, \tcg{mate}, \tcg{night}, \tcr{variety}, \tcb{ruminant}, \tcb{ridge}, \tcb{broad}, \tcb{scent}, \tcb{herd} \\

Beaver & \tcb{river}, \tcb{protect}, \tcb{semiaquatic}, \tcg{web}, \tcb{branch}, \tcb{eurasian}, \tcb{american}, \tcb{land}, \tcb{insular}, \tcg{hunt}, \tcb{fur}, \tcg{extant}, \tcr{adult}, \tcb{stream}, \tcb{pond} \\

Blue Whale & \tcb{ton}, \tcb{whale}, \tcb{flipper}, \tcg{kilometre}, \tcb{marine}, \tcb{ocean}, \tcr{belong}, \tcg{mph}, \tcg{shape},  \tcb{dive}, \tcg{earth}, \tcg{worldwide}, \tcb{indian}, \tcb{travel}, \tcb{pacific}\\

Buffalo & \tcg{climate}, \tcg{extant}, \tcb{herd}, \tcb{indian}, \tcb{cattle}, \tcb{dairy}, \tcr{animate}, \tcg{bc}, \tcb{trade},  \tcr{behaviour}, \tcg{human}, \tcb{milk}, \tcg{northern}, \tcg{southeast}, \tcb{field}\\

Cow & \tcb{draft}, \tcb{milk}, \tcb{cattle}, \tcb{widespread}, \tcb{product}, \tcb{meat}, \tcb{domestic}, \tcb{strong}, \tcb{cart},  \tcb{plow}, \tcb{oxen}, \tcb{bullock}, \tcb{cow}, \tcr{animate}, \tcb{india}\\

Deer & \tcb{antler}, \tcb{fight}, \tcg{mate}, \tcb{elk}, \tcb{palmate}, \tcb{moose}, \tcb{wolf}, \tcb{season}, \tcb{bear},  \tcb{woodland}, \tcb{herd}, \tcb{ruminant}, \tcb{deer}, \tcg{stomach}, \tcb{spike}\\

Moose & \tcb{herd}, \tcb{elk}, \tcb{palmate}, \tcb{moose}, \tcb{wolf}, \tcb{fight}, \tcb{deer}, \tcb{compete}, \tcb{alces},  \tcb{temperate}, \tcg{climate}, \tcb{aggressive}, \tcb{sedentary},  \tcg{season}\\

Mouse & \tcb{rodent}, \tcb{house}, \tcb{eat}, \tcg{avoid}, \tcb{burrow}, \tcr{general}, \tcr{genetic},  \tcb{popular}, \tcb{breed}, \tcb{wild}, \tcb{small},  \tcb{tail}, \tcb{vermin}, \tcb{nocturnal}, \tcb{prey}\\

Dolphin & \tcb{flipper}, \tcb{whale}, \tcb{ton}, \tcg{kilometre}, \tcb{indian}, \tcb{dive}, \tcg{mph}, \tcg{earth},  \tcg{shape}, \tcb{blubber}, \tcr{belong}, \tcb{marine}, \tcb{ocean}, \tcb{capture}, \tcg{prevent}\\

Horse & \tcb{draft}, \tcb{strong}, \tcb{milk}, \tcb{meat}, \tcb{ungulate}, \tcb{equip}, \tcb{widespread}, \tcr{loose},  \tcg{past}, \tcb{history}, \tcb{compete}, \tcb{endure}, \tcb{technique}, \tcg{style}, \tcb{flee}\\

Hamster & \tcb{mix}, \tcb{underground}, \tcb{fragile}, \tcb{house}, \tcb{bear}, \tcb{seed}, \tcr{worn},  \tcb{silky}, \tcb{rapid}, \tcg{classify}, \tcr{general},  \tcb{tail}, \tcb{flexible}, \tcb{dwarf}, \tcb{pouch}\\

Killer Whale & \tcb{ton}, \tcb{whale}, \tcb{dolphin}, \tcb{click}, \tcb{dive}, \tcb{killer}, \tcb{pollution},  \tcr{belong}, \tcb{capture}, \tcb{vocal}, \tcb{calf},  \tcb{tail}, \tcb{threat}, \tcb{fish}, \tcb{fin}\\

Otter & \tcb{semiaquatic}, \tcb{branch}, \tcb{eurasian}, \tcb{lake}, \tcr{engage}, \tcg{bed}, \tcb{play},  \tcb{trap}, \tcb{river}, \tcg{deplete}, \tcr{giant}, \tcb{cetacean}, \tcb{mink}, \tcb{weasel}, \tcg{web}\\

Rabbit & \tcb{fragile}, \tcb{house}, \tcg{classify}, \tcr{general}, \tcg{introduce}, \tcb{underground}, \tcb{pad},  \tcb{vegetarian}, \tcb{companionship},  \tcb{defensive}, \tcb{shelf}, \tcb{detect}\\

S. Monkey & \tcb{agile}, \tcb{arm}, \tcb{walk}, \tcb{tropic}, \tcb{rainforest}, \tcb{primate}, \tcr{source},  \tcg{primary}, \tcb{bark},  \tcr{passage}, \tcb{balance}, \tcb{thumb}, \tcb{moist}, \tcb{threaten}\\

\hline
\end{tabular}
\end{center}
\caption{Category-wisely top ranked words, sorted by average importance weights within each class. The \tcb{blue} words are generally considered as meaningful attributes of this class. The \tcg{green} words are concepts somewhat related to this class, but are less informative to define it. The \tcr{red} words are concepts that are not semantically related to the corresponding class.}
\label{tab: discAttrAwA}
\end{table*}

\section{Conclusion}
In this paper, we have introduced a noise suppression mechanism to text-based zero-shot learning. The proposed $l_{2,1}$-norm based objective function generates classifiers that are robust against textual noise and achieve state-of-the-art zero-shot learning performance. We have made several findings in the experiments. (1) The inherent noise within text sources has a significant impact on  zero-shot learning performance. As all the text-methods without noise suppression are inferior to our approach, we speculate that noise in a  component of the mid-level representation decreases its discriminative power. (2) Most noisy components are suppressed rather than completely eliminated by our mechanism. Some words, although unimportant individually, can produce meaningful discriminative power when put together.  (3) We find three kinds of words in the de-noised representation that can provide useful information for zero-shot learning. The first kind are the attribute-like words that explicitly describe the category. The second are words that are weakly related to the category. They usually occur with definitive words. The last kind of words is non-informative to humans, but shows certain distribution patterns among related categories.

Overall, this paper points out an important factor in text-based zero-shot learning that has been previously ignored. By
dealing directly with the inevitable variations in human expression, and suppressing words that contain little or no value, the performance of text-based automatic zero-shot learning can be significantly improved.

{\small
\bibliographystyle{ieee}
\bibliography{ref}
}

\end{document}